\documentclass[11pt,a4paper]{article}

\usepackage[utf8]{inputenc}
\usepackage[T1]{fontenc}
\usepackage{lmodern}
\usepackage[margin=1in]{geometry}
\usepackage{booktabs}
\usepackage{array}
\usepackage{amsmath}
\usepackage{microtype}
\usepackage[hidelinks]{hyperref}
\usepackage{xcolor}

\input{shared-preamble}

\title{A Policy Profile for Croissant:\\ Refusal as a Property of the Dataset}
\author{
  Alexander Chernov\\
  {\normalsize Independent Researcher}\\
  {\normalsize Toronto, Ontario, Canada}\\
  {\normalsize ORCID: \href{https://orcid.org/0009-0007-3198-2712}{0009-0007-3198-2712}}
}
\date{\today}

\begin{document}
\maketitle

\begin{abstract}
Croissant is the de facto machine-readable descriptor for ML datasets: JSON-LD
over schema.org. Since version 1.1 it also carries data use conditions,
recommending DUO and ODRL for them. What no version specifies is how any of them
is evaluated: no decision procedure, no bound on evaluation cost, no outcome for
a condition an implementation cannot evaluate, no record of what was checked,
and nothing on composition with caller-side authority.

We supply that half. An additive profile lets a dataset declare the operations
it admits and the conditions under which it admits them, over a closed set of
five operators whose decision procedure is given in full, so a gate decides from
the descriptor alone and records what it checked.

Two corpora evaluate it and their evidence is kept apart. Three descriptors that
gated a real nf-core pipeline give the deployment result: decisions from a
profile document match the gate's native descriptor record for record, stripping
the layer leaves a valid Croissant document, and the added cost is $11.7\us$
against a $119\us$ decision. A corpus generated from the profile's grammar gives
the breadth, covering every operator, refusal class and conformance clause.
Across its valid cases, 552 complete decision records agree three ways---native
descriptor, profile terms, and the same policy as ODRL in \code{usageInfo}. The
carrier is therefore not the contribution; the evaluation semantics is.

Finally, caller-bound and data-bound policies range over non-overlapping state
spaces, so neither permit set contains the other.
\end{abstract}


\section{The gap}

Three threads are moving on dataset descriptors and only their intersection is
empty.

The \textbf{venue thread} treats the dataset as a citable object, in the
tradition the FAIR principles set out~\cite{wilkinson2016fair}: \emph{Scientific
Data} pioneered the Data Descriptor article type, Frontiers' FAIR\textsuperscript{2}
articles add AI-readiness and executable notebooks, and in 2026 the \emph{Journal
of Biomedical Physics and Engineering} adopted the type as
well~\cite{mahmoudi2026datadescriptors}. The
\textbf{format thread} is Croissant~\cite{akhtar2024croissant,akhtar2024croissantneurips}:
JSON-LD~\cite{kellogg2020jsonld} over schema.org, now the interoperable
serialisation across the major dataset hubs---published by Hugging
Face~\cite{hf2024croissant}, Kaggle and OpenML, and consumed by Google Dataset
Search~\cite{google2026datasetsearch}. The \textbf{actionability thread}
is Croissant over MCP~\cite{anthropic2024mcp}: an endpoint through which an
agent discovers, downloads and loads a dataset.

Until this year the gap was one of representation, and it is not any more.
Croissant 1.1~\cite{mlcommons2026croissant11}, published in January 2026, adds a
responsible-AI and governance section that carries data use conditions:
restrictions travel in \code{sc:usageInfo}, DUO~\cite{lawson2021duo} is
recommended for simple conditions and ODRL~\cite{iannella2018odrl} for
fine-grained ones, and the specification states that machine-readable use
restrictions can support automated compliance checking.

What it does not state is how. A vocabulary in which a condition can be written
is not a procedure by which a request is decided, and Croissant 1.1 specifies no
evaluation semantics for the conditions it now carries, no failure semantics for
a condition an evaluator does not implement, no record of what was checked, and
nothing about how a condition on the data composes with the authority that
governs the caller. All three threads describe, and the third now describes
conditions of use. None of them refuses. A descriptor that an agent can act on,
in a governed setting, has to be able to say no---and to say what it checked when
it did.

\paragraph{Where this sits.} The wider framing is the \emph{agentic
dataset}~\cite{chernov2026agenticdatasets}: a dataset treated as a participant in
an engineering control plane rather than as a passive object an external system
decides about. That work argues the position; this paper is a narrow, checkable
instance of it, and deliberately the smallest one we could find. A dataset becomes
agentic here in exactly one respect---it carries the conditions of its own use in
a standard descriptor, so a gate can decide from the descriptor alone---and in no
other. It gains no autonomy, no identity model, no ability to initiate anything.
Whether that minimal form is worth having is the question the evaluation answers,
and it is a much smaller question than the framing paper's.

\paragraph{Contributions.} The contribution is the executable half, not the
vocabulary half. This paper contributes:

\begin{enumerate}
\item An additive policy profile for Croissant with a \emph{specified evaluation
semantics}---a total decision procedure for every operator, stated in the paper
(Table~\ref{tab:operators}) rather than left to the artifact.
\item A closed operator language whose consequences are the point rather than
the price: evaluation cost enumerable from the document, a fail-closed outcome
for anything the evaluator cannot check, and a deterministic projection onto a
machine-readable capability schema, so an advertised constraint and an enforced
one have one source.
\item Two carriers that decide identically---the profile's own terms, and an
ODRL policy in the \code{sc:usageInfo} slot Croissant 1.1 provides---which
settles by demonstration that the carrier is not the contribution.
\item A two-part evaluation that keeps its evidence separate: a deployment
corpus of descriptors that gated a real bioinformatics pipeline, which is where
every timing figure comes from, and a corpus generated from the profile's
grammar, which covers every operator, refusal class and conformance clause and
which found two defects the first could not reach.
\item A structural result on composing caller-side and data-side authority,
witnessed on the deployment corpus: neither permit set contains the other,
because the two range over state spaces that do not overlap.
\end{enumerate}

The profile is published with the artifacts a consumer needs to act on it
without reading this paper: a specification, a JSON-LD context, SHACL shapes for
the static conformance clauses, both carriers, and a description in the Profiles
Vocabulary saying which artifact plays which role.

Section~\ref{sec:related} places the profile among the vocabularies it extends
and the policy languages it declines to be, including the three questions a
reader from this field asks first: whether Croissant 1.1 already subsumes it,
why not an ODRL profile, and why not SHACL.
Section~\ref{sec:design} gives the design, Section~\ref{sec:eval} the
measurements, and Section~\ref{sec:precedence} the precedence result.

\section{Related work}
\label{sec:related}

\subsection{Descriptor vocabularies and the additive profile}

Croissant is JSON-LD~\cite{kellogg2020jsonld} over
schema.org~\cite{guha2016schemaorg}, and it sits inside a settled pattern for
extending such a vocabulary without forking it.
DCAT~\cite{albertoni2024dcat3} is profiled by DCAT-AP~\cite{semic2025dcatap} for
European data portals, and DCAT-AP in turn by
MLDCAT-AP~\cite{semic2026mldcatap} for machine-learning models and the datasets
they were trained on. The Profiles Vocabulary~\cite{atkinson2019prof} supplies
the terms in which a profile describes itself and the resources---a
specification, a context, a validator---that make it usable by someone who
encounters it. The layer defined here is that pattern applied to Croissant, and
Section~\ref{sec:layer} states the additivity condition it holds itself to.

What none of these carry is a decision. MLDCAT-AP describes a model, its
datasets and the quality measured on them; DCAT-AP describes a distribution and
attaches its licence as a link to a document meant for a person. Croissant's own
responsible-AI extension~\cite{mlcommons2024raispec,jain2024croissantrai} comes
closest, and is worth being precise about because it is the neighbour a reader
reaches for first. Croissant-RAI adds terms for the data lifecycle, labelling
protocols, known limitations and biases, sensitive categories, intended use
cases and compliance; the reference implementation of this profile emits several
of them, through the \code{rai:} prefix, alongside its own. They are
documentation fields. They record what someone auditing a dataset should know,
and no term among them takes a request and returns permit or refuse. The profile
defined here is additive to Croissant-RAI in the same sense that it is additive
to Croissant, and answers a question Croissant-RAI does not pose.

That Croissant is a governance instrument and not only a loading format is
already argued in this journal: KG.GOV~\cite{meronopenuela2025kggov} takes it as
one of three use cases for knowledge graphs as the backbone of data governance in
AI. We take that position as given and ask what has to be added to make it
executable.

\subsection{Croissant 1.1 data use conditions}
\label{sec:croissant11}

Croissant 1.1~\cite{mlcommons2026croissant11} is the nearest thing to a
subsuming baseline and has to be dealt with directly. Its responsible-AI and
governance section carries data use conditions in \code{sc:usageInfo}, points at
DUO~\cite{lawson2021duo} for simple conditions---a GA4GH standard vocabulary of
consent and data-use terms already applied to a large number of biomedical
datasets---and at ODRL~\cite{iannella2018odrl} for hierarchical permissions and
modifiers. It says that machine-readable use restrictions can support automated
compliance checking. On representation, the question this paper opened is
answered, and by the standard rather than by us.

Five things separate representing a condition from deciding a request, and
Croissant 1.1 supplies none of them.

\emph{Evaluation.} DUO terms are a vocabulary of permissions and restrictions.
ODRL is a model that leaves the evaluator to the implementer
(Section~\ref{sec:policylang}). Neither Croissant 1.1 nor either of the
vocabularies it recommends says what it means to check a condition against a
request, so two conforming consumers may reach different verdicts on the same
document without either being wrong.

\emph{Bounds.} Nothing in \code{usageInfo} constrains what a condition may cost
to evaluate. A closed operator set makes the work a document implies enumerable
from the document, and that is a property of the profile, not of the carrier.

\emph{Failure semantics.} A condition an implementation does not understand has
no specified outcome. Whether the request is refused or the condition is skipped
decides whether the mechanism is a gate or a suggestion, and the specification
does not say.

\emph{Record.} A compliance check that leaves no record of the conditions it
evaluated, with the values it observed, cannot be audited after the fact and
cannot be re-decided against a later policy. Section~\ref{sec:eval} treats the
decision record as the unit of comparison for exactly this reason.

\emph{Composition.} A dataset-side condition and a caller-side authority both
constrain the same request. Croissant 1.1 does not address the relationship, and
Section~\ref{sec:precedence} shows that ignoring it admits requests one authority
would have refused, in both directions.

The consequence for this paper is that its claim is not that Croissant cannot
express policy. It is that expressing policy and admitting a request are
different problems, and that the second is where the properties an auditor needs
are won or lost. Croissant 1.1 makes that the more interesting claim rather than
a weaker one.

\subsection{Policy languages on the Web}
\label{sec:policylang}

ODRL~\cite{iannella2018odrl} is the Recommendation for expressing permissions,
prohibitions and duties over assets, and it is deliberately a model rather than
an engine: it defines what a policy says and leaves to the implementer how a
policy is decided. Supplying that second half is an active line of work.
De Vos et al.~\cite{devos2019odrl} translate ODRL into Answer Set Programming so
that compliance can be checked against regulatory requirements, and
ODRE~\cite{cimmino2025odre} attaches executable functions to constraints so that
a policy can be enforced rather than only read. That two independent groups
supply two different evaluation semantics for the same standard is the fact
Section~\ref{sec:whynotodrl} turns on.

The Data Privacy Vocabulary~\cite{pandit2019dpvcg,dpvcg2025dpv} is the other
large vocabulary in this space and the closest in intent to what is defined
here: it states what may be done with data, in RDF, in terms a machine reads. Its
subject is personal-data processing---purposes, legal bases, processing
operations, recipients---and, like ODRL, it is a vocabulary rather than a
decision procedure.

Outside the Web stack, XACML~\cite{oasis2013xacml} has had both halves since the
early 2000s: an attribute-based model and a specified evaluation with named
rule- and policy-combining algorithms. It is the standard this profile most
resembles in what it wants and least resembles in how that is carried. XACML
policies live in a policy store beside a decision point. They do not travel in
the descriptor a data hub already serves, which is the property
Section~\ref{sec:precedence} shows is load-bearing.

\subsection{Why not an ODRL profile}
\label{sec:whynotodrl}

ODRL has a profile mechanism, and it is more than adequate for what is expressed
here: Section~3.3 of the Information Model~\cite{iannella2018odrl} lets a
profile create new actions, new left and right operands, new operators and new
conflict strategies. A dataset's lifecycle state would be a left operand,
\code{QC\_PASSED} a right operand, and \code{min} and \code{max} are in the
common vocabulary already. The expressiveness objection to ODRL does not hold, and we do not make
it.

Three things an ODRL profile does not give, and each is a reason this layer
exists.

\emph{Bounded evaluation is not inherited.} Creating an operator in a profile
gives it an IRI and a definition in prose. It does not give it a decision
procedure, because the standard does not specify one---which is precisely what
\cite{devos2019odrl} and \cite{cimmino2025odre} independently supply, and they
supply different things. The property this work needs is that the work a document
implies is enumerable from the document. No profile of a standard that leaves
evaluation open can promise that; the promise has to come from closing the
operator set and naming the evaluator, which is what Section~\ref{sec:closed}
does.

\emph{Halting at profile granularity is not refusal at condition granularity.}
ODRL is stricter on this point than it is usually given credit for. If an ODRL
processing system does not recognise a profile identifier it \emph{must} stop
processing the policy (Information Model, Section~3.2). Failing closed against an
unknown profile is normative, and we do not argue otherwise.

Two questions sit beneath that requirement, and a gate has to answer both. The
standard does not say what a processor does with an operator it does not
implement inside a profile it \emph{does} recognise---a live case precisely
because the profile mechanism exists to add operators. And it does not say what
the surrounding system concludes from a halted evaluation: whether the absence of
a decision is a refusal or merely the absence of an answer. Both are properties
of a deployment rather than of the policy language. Closing the operator set
removes the first question, because within a recognised version there is no
unrecognised operator. Delegating to an existing gate answers the second, because
that gate already maps a condition it cannot evaluate onto a refusal it can
(Section~\ref{sec:failclosed}).

\emph{Nothing projects.} The capability schemas of
Section~\ref{sec:projection} are generated because each of the five operators
has an image in JSON Schema: \code{min} becomes \code{minimum}, \code{in}
becomes an \code{enum}. An operator minted by a profile has no such image, and a
constraint carrying an executable function has none in principle. The projection
is a property of the closed set rather than a feature added beside it.

\paragraph{We built it anyway.} None of the three is an argument against
expressing this policy in ODRL, and Croissant 1.1 names ODRL in
\code{sc:usageInfo} as the place to put it. So the profile also defines an ODRL
carrier: the same policy as an \code{odrl:Set} in \code{usageInfo}, under a
distinct profile identifier, translated by the same code into the same native
model and decided by the same gate. Section~\ref{sec:c5} reports what that
settles.

Two things are worth stating here because they bear on the argument rather than
on the measurement. First, \textbf{four of the five operators are ODRL core
operators}---\code{min}, \code{max}, \code{in} and \code{equals} are
\code{odrl:gteq}, \code{odrl:lteq}, \code{odrl:isAnyOf} and \code{odrl:eq}. Only
the presence test is minted, because ODRL defines no existence operator. That is
the shortest available statement of the position taken above: the disagreement
was never about expressiveness, and a paper claiming otherwise would be wrong on
a checkable point.

Second, the carrier gets its fail-closed behaviour against an unknown profile
from ODRL rather than from us, and the implementation does exactly what
Section~3.2 requires: a policy whose \code{odrl:profile} the reader does not
recognise yields a descriptor with no admissible action, so every request is
refused with the gate's own record rather than raising an exception a caller
might catch and continue past.

One remaining difference is of degree only, and we rest nothing on it. An ODRL
policy is a separate graph linked to its target, though it may be serialised in
the same JSON-LD document; the \code{cpol:} terms hang off the \code{sc:Dataset}
node, so a consumer that already parses Croissant meets the policy without a
second vocabulary stack.

\subsection{Why not SHACL}
\label{sec:whynotshacl}

The five operators look like a small fragment of
SHACL~\cite{knublauch2017shacl}---\code{sh:minInclusive}, \code{sh:maxInclusive},
\code{sh:in}, \code{sh:hasValue}, \code{sh:minCount} cover them---and the
question is fair. The mismatch is not in the vocabulary but in what is being
constrained. SHACL validates a data graph that exists and returns a validation
report. The object of a decision here is a request context supplied at call time,
which is in no graph until it is put in one: expressing the check would mean
injecting each request into the data graph and validating per request, and the
result would be a report about a graph rather than a record about a request. The
second problem is the same one as before. The fragment that stays bounded is a
fragment, and SHACL-SPARQL constraints are as expressive as the query language,
so ``conforms to SHACL'' carries no bound on evaluation at all.

Where SHACL does fit is the conformance checking of the profile documents
themselves, and that is not a hypothetical: the profile publishes SHACL shapes
for its static conformance clauses, and Section~\ref{sec:shapes} reports what
running them found. The division is clean and worth keeping in view for the rest
of the paper. \textbf{SHACL validates the policy document. The gate decides the
request.} A document can satisfy every shape and refuse everything it is asked,
and that is not a contradiction---the shapes constrain how a policy is written,
not what it permits.

\subsection{Policies attached to Web resources}

Associating a policy with a Web resource and reasoning over it with Semantic Web
technology is not new, and the lineage matters because the surface resemblance is
close. Rein~\cite{kagal2006rein} places policies at URIs alongside the resources
they govern and reasons over them across rule languages, so that policy is as
decentralised as the Web it governs. Solid's Access Control
Policy~\cite{bosquet2022acp} is a current form of the same idea: an RDF policy
language in which a request context---agent, client application, issuer---is
matched against authorization graphs attached to a resource to determine which
access modes are granted.

Both of these decide, which is more than can be said for the vocabularies above,
and neither is subsumed by what is defined here. Nor does either subsume it,
because the predicate differs. Rein and ACP decide access to a resource, and what
they range over is principally the requester: who is asking, on whose authority,
through which client. The layer defined here decides an operation against the
scientific and operational state of the dataset itself---whether the data has
passed QC, whether a required measurement is present, whether a threshold is
met---and it has no identity model at all, which Section~\ref{sec:limits} states
as a limitation rather than an omission. A request that ACP admits and this
profile refuses is not a contradiction. The two are answering different
questions, which is the same structure Section~\ref{sec:precedence} turns into a
measurement.

\subsection{Recording, and deciding elsewhere}

Two further bodies of work sit close enough to be mistaken for this one, and
neither closes the gap.

RO-Crate~\cite{soilandreyes2022rocrate} and Workflow Run
RO-Crate~\cite{leo2024workflowrun} package research artifacts and record what a
workflow did, in RDF, with the same commitment to travelling with the data that
motivates the layer here. They record retrospectively. A run that should not have
happened is described exactly as faithfully as one that should, because the
record is written after the fact and nothing consults it before.

Agent control planes---watsonx Orchestrate's Agentic Control
Plane~\cite{ibm2026controlplane} and Drata's AI Agent
Governance~\cite{drata2026agentgov}---do decide, and inline: they evaluate an
action against approved policy before execution and block violations. That is
structurally the same mechanism as the gate below, aimed at a different subject,
because they bind policy to the \emph{caller}. A rule about what a dataset admits
cannot be bound to a caller, because it does not travel with one. What follows
from that is not an opinion about which binding is right, and
Section~\ref{sec:precedence} makes it a measurement instead.

\subsection{Combining two authorities is old, and we do not claim it}

The conjunction of Section~\ref{sec:precedence} is a mechanism with a long
record, and we want it clear at the outset that we are not claiming it.
XACML~\cite{oasis2013xacml} has named rule- and policy-combining algorithms
since the early 2000s---\code{deny-overrides} among them, by that name---and its
ABAC model evaluates subject-side and resource-side attributes within a single
decision. Oracle's US10230732B2~\cite{vepa2019policyobjects} (priority 2013)
combines policies from separately administered sources under a selected
combining algorithm, deny-override by default. AWS Clean
Rooms~\cite{aws2023cleanrooms} enforces the most restrictive control across
per-collaborator rules whose authors cannot see each other's data, and IAM
cross-account access requires a resource-owner policy and a caller-account policy
to agree. The mechanism of Section~\ref{sec:precedence} is therefore not new, and
any claim that consulting two authorities is unspecified would be wrong.

Two things remain. First, these systems bind the resource-side half to a
\emph{storage or account boundary}: a condition on the lifecycle state of a
dataset---\code{QC\_PASSED}, \code{PENDING}---is not expressible in any of them,
because the vocabulary is access to an object rather than the scientific state of
its contents. Second, none of them travels: the policy lives in one provider's
control plane, not in a descriptor that moves with the data across organisations.
What Section~\ref{sec:precedence} contributes is not the conjunction but a
measurement of what is lost when the data-side half is absent, which is the
condition of every deployed agent control plane we are aware of.

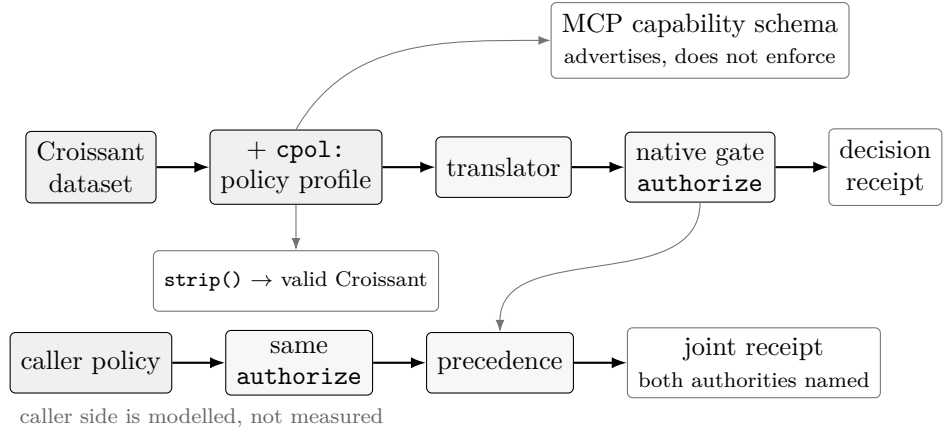
\begin{figure*}[t]
\centering
\begin{tikzpicture}[
  font=\small,
  box/.style={draw, rounded corners=2pt, align=center, inner sep=4pt,
              minimum height=8mm, fill=black!3},
  doc/.style={box, fill=black!6},
  result/.style={box, fill=white, draw=black!55},
  flow/.style={-{Latex[length=2mm]}, thick},
  sideflow/.style={-{Latex[length=1.6mm]}, draw=black!55},
]

\node[doc] (cr) {Croissant\\ dataset};
\node[doc, right=7mm of cr] (prof) {{}+ \code{cpol:}\\ policy profile};
\node[box, right=7mm of prof] (tr) {translator};
\node[box, right=7mm of tr] (gate) {native gate\\ \code{authorize}};
\node[result, right=7mm of gate] (rec) {decision\\ receipt};

\draw[flow] (cr) -- (prof);
\draw[flow] (prof) -- (tr);
\draw[flow] (tr) -- (gate);
\draw[flow] (gate) -- (rec);

\node[result, above=7mm of gate] (mcp) {MCP capability schema\\ \scriptsize advertises, does not enforce};
\draw[sideflow] (prof.north) to[out=60, in=180] (mcp.west);

\node[result, below=6mm of prof] (strip) {\scriptsize \code{strip()} $\rightarrow$ valid Croissant};
\draw[sideflow] (prof.south) -- (strip.north);

\node[doc, below=17mm of cr] (caller) {caller policy};
\node[box, right=7mm of caller] (g2) {same\\ \code{authorize}};
\node[box, right=7mm of g2] (prec) {precedence};
\node[result, right=7mm of prec] (joint) {joint receipt\\ \scriptsize both authorities named};

\draw[flow] (caller) -- (g2);
\draw[flow] (g2) -- (prec);
\draw[flow] (prec) -- (joint);
\draw[sideflow] (gate.south) to[out=270, in=90] (prec.north);

\node[anchor=west, font=\scriptsize, text=black!60] at ([yshift=-3.2mm]caller.south west)
  {caller side is modelled, not measured};

\end{tikzpicture}
\caption{The profile is additive: \code{cpol:} terms hang off an ordinary
Croissant dataset, a translator maps them to the native descriptor model, and the
decision is taken by a gate that already existed. Two branches come off that
spine. The capability projection is generated from the same document the gate
enforces, so the advertised constraint and the enforced one are not transcribed
apart by hand; it advertises and never replaces the check. Stripping the
\code{cpol:} terms leaves a document a profile-unaware consumer still loads.
Below, the same evaluator decides a caller-side authority, and a precedence rule
combines the two into one receipt naming both.}
\label{fig:arch}
\end{figure*}

\section{Design}
\label{sec:design}

\subsection{One constraint}

A descriptor that can be refused from is only useful if the refusal is cheap,
total, and auditable. Cheap, because a gate on every access must not be the thing
you profile. Total, because a rule the enforcer does not understand must stop the
work rather than be skipped. Auditable, because a refusal that cannot say what
was checked is indistinguishable from a bug.

\subsection{The layer}
\label{sec:layer}

A \code{cpol:Policy} attaches to the Croissant \code{sc:Dataset} node. It carries
the dataset's lifecycle state, a mandatory \code{failClosed}, informative
handling metadata, and one or more \code{cpol:Action} nodes. Each action names
the states that admit it and a list of \code{cpol:Condition} nodes evaluated
against the request context.

The profile is additive by construction: removing every \code{cpol:}-prefixed
term must leave a valid Croissant document, and the layer must not redefine any
core term. Both halves of that condition are properties of the JSON-LD
context~\cite{kellogg2020jsonld}. The \code{@context} adds exactly two
entries---the namespace prefix, and one typed term for the condition
operand---and every policy key is written prefixed rather than aliased to a bare
name, paying the verbosity to keep the guarantee: an aliased term is
indistinguishable from a core term after expansion, and a stripped document
would have to be re-checked to know whether anything had been redefined.

The profile is identified by a versioned IRI and declared through
\code{conformsTo}, and it publishes the three resources a consumer needs to act
on that declaration---a specification document, the JSON-LD context, and a
conformance validator. Section~\ref{sec:threats} records where those resources
dereference. Those are described in the terms of the Profiles
Vocabulary~\cite{atkinson2019prof}: the profile publishes a \code{prof:Profile}
naming what it is a profile of and a \code{prof:ResourceDescriptor} for each
artifact, with roles drawn from PROF's own eight rather than invented. A
consumer holding a document that claims the profile can therefore find the
validator from the identifier instead of from this paper. The ODRL carrier of
Section~\ref{sec:whynotodrl} is described there too, as a profile of both
Croissant and ODRL and under its own identifier, because a processor required to
stop at an identifier it does not recognise gains nothing if the two carriers
share one.

\subsection{The closed operator set}
\label{sec:closed}

\code{min}, \code{max}, \code{in}, \code{equals}, \code{present}. That is the
whole language, and because the paper's central claim is that the profile
supplies an evaluation semantics the carrier does not, the semantics belongs in
the paper rather than only in the artifact.

\begin{table*}[tp]
\centering
\footnotesize
\setlength{\tabcolsep}{4pt}
\begin{tabular}{@{}llll@{}}
\toprule
operator & decision, observed $o$ vs.\ operand $e$ & ODRL & JSON Schema \\
\midrule
\code{min}     & $o \ge e$; refuse if either is non-numeric & \code{odrl:gteq}   & \code{minimum} \\[2pt]
\code{max}     & $o \le e$; refuse if either is non-numeric & \code{odrl:lteq}   & \code{maximum} \\[2pt]
\code{in}      & $o$ equals a member of the array $e$       & \code{odrl:isAnyOf}& \code{enum} \\[2pt]
\code{equals}  & $o = e$                                    & \code{odrl:eq}     & \code{const} \\[2pt]
\code{present} & $(o \ne \bot) = e$, $e$ boolean            & minted             & \code{required}, negated \\
\bottomrule
\end{tabular}
\caption{The whole language. Each row is a total function from an observed value
to a verdict, for a \emph{well-typed} operand, and that qualifier is where the
totality claim is earned rather than assumed. The operand is written in the
document and the observation arrives with the request, and they fail
differently. An observation of the wrong type is an ordinary condition
violation---\code{min} against a non-numeric observation refuses rather than
coercing, and a missing observation is $\bot$ rather than an exception. An
operand of the wrong type is not a request problem at all but a defect in the
policy, and it is refused by the translation of Section~\ref{sec:failclosed}
before any comparison is attempted: \code{min} and \code{max} require a number,
\code{in} a JSON array, \code{present} a boolean, and \code{equals} any JSON
value. \code{equals} and \code{in} share one comparison relation, which the
profile defines rather than inherits: numbers compare by value, so $1$ and $1.0$ are
equal; a boolean is equal only to a boolean, so \texttt{true} is not $1$; arrays
and objects compare structurally under the same relation, and duplicates in an
\code{in} array do not change membership; values of different JSON types are
unequal. Section~\ref{sec:conformance} reports what happened when these
boundaries were left implicit. The third column is the ODRL carrier of
Section~\ref{sec:whynotodrl}---four rows are ODRL core operators and only
\code{present} is minted, because ODRL defines no existence operator. The fourth
is the capability projection of Section~\ref{sec:projection}, which exists
because every row has an image in JSON Schema; \code{present: false} projects as
the negation of \code{required}, since the constraint is that the property be
absent. An operator outside these five has no row, and an evaluator that meets
one refuses.}
\label{tab:operators}
\end{table*}

The case against a larger set is made in Section~\ref{sec:whynotodrl} and not
repeated here. What closing it buys is stated positively: five operators mean
each primitive condition has bounded, explicitly specified evaluation semantics,
and a decision record can state in one line what was compared to what.
The cost of a whole decision is linear in the number of conditions checked: there
is no recursive policy language, no rule that can invoke another, and no
unbounded evaluation anywhere in the profile. We claim that and not
constant-time---evaluating $n$ conditions is $O(n)$, and \code{in} carries a
further term in the cardinality of its allowed set unless that set is
hashed---because the bound that matters to an auditor is that the work is
enumerable from the document, not that it is $O(1)$. Extending the set is a version bump
rather than a profile extension, because an evaluator must refuse an operator it
does not implement: a sixth operator added silently to a document turns PERMIT
into REFUSE, never into an unchecked pass.

\subsection{Fail-closed as a property of the translation}
\label{sec:failclosed}

The profile introduces no separate evaluator. It defines the decision procedure
of Table~\ref{tab:operators} and implements it by translating a document into
the native descriptor model of a gate that already exists and calling that gate.
The semantics is the profile's; the engine is reused.

This is what makes fail-closed cheap to believe rather than cheap to claim. The
native evaluator already refuses on a condition whose operator it does not
recognise. The translation maps every defect---unknown operator, missing field,
duplicate condition name, a policy that does not declare \code{failClosed}---onto
exactly that condition.

The profile therefore introduces \emph{no second decision engine}: enforcement
stays delegated to the existing gate, and what the profile adds is a translation.
That is not the same as adding nothing that can be wrong. It relocates the
correctness obligation onto translation fidelity, which is precisely what
C1 (Section~\ref{sec:c1}) exists to test. What the design does buy is the failure
mode: a translation that fails produces a refusal rather than a gap.

\subsection{Capability projection}
\label{sec:projection}

Each action projects onto one MCP tool: a name, a generated description naming
the required states, and an input schema with one property per condition, typed
and constrained from the operator (\code{min} becomes \code{minimum}, \code{in}
becomes an \code{enum}, and so on).

Two consequences are intended. The constraint appears in the schema, so a
well-behaved caller can avoid a refusal instead of discovering it. And the schema
is derived rather than authored, so the advertised constraint and the enforced
one are generated from a single source and cannot be transcribed apart.
Hand-transcribed registry entries are the failure mode this removes.

The claim is single-source derivation, not that drift becomes impossible. A stale
registry entry, a cached schema, a generator-version difference or deployment
skew between the component that published a schema and the one now enforcing it
all remain possible. What is eliminated is the drift introduced by a human
copying a threshold from one artifact into another.

The projection describes what will be checked and never replaces the check. A
caller that satisfies the schema is still gated, which the evaluation of a state
precondition demonstrates directly: state is a property of the dataset, the
schema cannot carry it, and a request satisfying every advertised constraint is
still refused when the state is wrong.

\section{Evaluation}
\label{sec:eval}

\subsection{Two corpora}
\label{sec:corpora}

Two corpora answer two questions, and keeping them apart is what lets either
claim stand. The \textbf{deployment corpus} is three descriptors that gated a
real pipeline run; it is the evidence that the mechanism works in front of real
execution, and every timing figure below comes from it. The \textbf{conformance
corpus} of Section~\ref{sec:conformance} is generated from the profile's
grammar; it is the evidence that the profile behaves correctly across every
feature the specification defines, and it carries no timing at all. Three real
descriptors are thin evidence about a feature space, and generated documents are
no evidence at all about production reality. Merging them would overstate the
second and waste the first.

\subsection{The deployment corpus}
\label{sec:deployment}

The deployment corpus is three descriptors that gated a real run of
\code{nf-core/demo} v1.0.1~\cite{ewels2020nfcore}---the FASTQC, SEQTK\_TRIM and
MULTIQC stages over Illumina amplicon test reads---through
Nextflow's~\cite{ditommaso2017nextflow}
\code{process.beforeScript} hook, where a non-zero exit stops the task before its
script runs. They are the descriptors of a prospective admission gate measured on
that pipeline at $119\us$ per decision, median over 210 decisions in 30
interleaved replicates, not fixtures written for this paper.

The gate, the corpus and the replication it comes from are archived
separately~\cite{chernov2026corpus}; the profile, its reference implementation
and the benchmarks reported here are archived at~\cite{chernov2026profile}. Both
are public, so every figure below is recomputable rather than reported.

\paragraph{C1, translation fidelity.}\label{sec:c1}
For every dataset, a request matrix is generated from the descriptor rather than
listed by hand: the satisfying context, the empty context, a context of
irrelevant keys, one violation per condition, a non-numeric value for each
numeric condition, and two undeclared action names. Each request is decided
twice---once from the native descriptor, once from the emitted profile
document---and the two complete decision records are compared, not merely the
verdicts. Two evaluators that agree on PERMIT/REFUSE and disagree on the refusal
class, the reasons, or the conditions they claim to have checked are not
equivalent in any sense an auditor would accept. All records match, and the
matrix is separately asserted to exercise PERMIT and all three refusal classes,
so that the agreement is not established over a set of cases that happens to be
trivial.

\textbf{What this does and does not establish.} Both sides of the comparison are
decided by the same native gate, deliberately---a second evaluator would let a
disagreement come from the evaluators rather than from the documents. The
consequence is that C1 establishes the \emph{fidelity of the Croissant-to-native
translation} and nothing about whether the gate's own authorization semantics are
correct. Those are assumed, and they are the subject of the artifact this profile
imports rather than of this paper. Where we write that decisions are equivalent,
the scope is always translation fidelity in that sense.

\paragraph{C2, graceful degradation.}
Stripping the layer leaves an \code{sc:Dataset} with its name, description,
distribution, provenance, custodian and schema intact and no residue of the
profile. The descriptive half of the native descriptor is carried in standard
vocabulary---\code{additionalType}, \code{isBasedOn}, \code{measurementTechnique},
\code{variableMeasured}, and schema.org's own \code{additionalProperty} escape
hatch for anything unmapped---rather than in \code{cpol:} terms, so a consumer
that ignores the profile loses the policy and nothing else. The \code{@context}
is checked term by term against Croissant's to confirm nothing is redefined.

Writing this test changed the specification. The profile IRI appears in
\code{conformsTo} as a \emph{value}, not as a \code{cpol:}-prefixed key, so a
literal reading of clause 1 leaves a stripped document still claiming conformance
to a profile whose terms are gone---a document that would fail the profile's own
clause 3. The strip now removes the claim with the terms, and the specification
says so.

\paragraph{C3, cost.}
5000 iterations per case, in-process, over the three descriptors, for a permitted
and a refused request each:

\begin{center}
\begin{tabular}{@{}p{0.46\linewidth}rrr@{}}
\toprule
regime & native & profile & added \\
\midrule
warm --- document translated once, descriptor reused & $1.8\us$ & $1.8\us$ & $0.0\us$ \\[2pt]
cold --- document read and translated per decision & $8.5\us$ & $20.2\us$ & $+11.7\us$ \\
\bottomrule
\end{tabular}
\end{center}

Warm is zero because it is the same function on the same object. Cold is where
the profile costs something, and most of that cost is not the profile: JSON
parsing accounts for about $9.7$ of the $11.7\us$, because the Croissant document
with its context is 3.1\,KB against the native descriptor's 754\,B. The
profile-specific translation is about $2\us$.

Projected onto the measured figure, the gate process that took $119\us$ end to
end becomes about $131\us$. Whether that is observable is worth stating
carefully, because the same harness answers it directly and the answer is not
the comfortable one. Replicating the gated and ungated arms 30 times each and
pairing by replicate, the per-task end-to-end cost of gating \emph{does} resolve:
$+25.7$\,ms, 95\% CI $[+3.3, +48.2]$. It is not the policy. It is one Python
interpreter spawned per task, measured separately at $30.2$\,ms against
$9.6$\,ms for a bare interpreter, a figure that falls inside that interval.

So the honest statement is a ratio rather than an absence. The decision costs
$119\us$ and the mechanism delivering it costs $30$\,ms---a factor of roughly
250---and the profile adds $11.7\us$ to the former. Expressing policy as a
standard descriptor is therefore invisible inside a per-task cost that is
dominated by process startup, and would remain small against a resident gate
that removed that startup entirely. An earlier single-run version of this
measurement reported the overhead as lying below the resolution of the engine's
own trace; at $n=30$ that is not true, and the claim is withdrawn here rather
than repeated.

\paragraph{C4, no drift.}
Changing a threshold in the document changes the advertised \code{minimum} and
the gate's verdict in the same edit, because both are derived from the same node.
An operator outside the closed set projects as an unsatisfiable parameter,
because a capability that can never be called should not advertise a callable
schema.

\subsection{Carrier independence}
\label{sec:c5}

\paragraph{C5.}
The same policy is emitted twice: once as \code{cpol:} terms on the
\code{sc:Dataset} node, once as an \code{odrl:Set} in \code{sc:usageInfo}, the
place Croissant 1.1 puts use conditions. Both are decided through the same
translation into the same native model by the same gate, and the whole decision
records are compared over the generated request matrix. They match: verdict,
refusal class, reasons, and every condition with its observed value.

Two controls make that worth reporting. The two documents are required to be
identical outside the policy node, so the comparison isolates the carrier rather
than comparing two documents somebody wrote separately; and the matrix is
separately asserted to exercise \code{PERMIT} and all three refusal classes on
the ODRL side, so agreement is not established over a set of cases that happens
to be trivial. Flipping one entry in the operator table---mapping \code{min} to
\code{odrl:lteq}---breaks nineteen of the comparisons, which is the check that
the test can fail at all.

What C5 establishes is narrow and, for the question of Section~\ref{sec:whynotodrl},
decisive. It is not that ODRL is unnecessary. It is that \emph{the carrier
decides nothing}: the evaluation semantics does, it is the same in both, and it
is what neither Croissant 1.1 nor ODRL supplies. The \code{cpol:} form is
normative here because it is smaller and because a Croissant consumer meets it
without a second vocabulary stack, and nothing in the paper rests on that choice.

\subsection{Conformance checking}
\label{sec:shapes}

\paragraph{Shapes, and what they found.}
The static conformance clauses are published as SHACL
shapes~\cite{knublauch2017shacl} over the expanded graph, which is the check the
profile's structural validator cannot perform---it inspects JSON and does not
parse JSON-LD. All three documents conform. Eight deliberately broken documents,
one per clause plus four structural defects, are each reported as violations,
because shapes that accept everything pass silently.

Writing them found a defect in Croissant, and it is worth reporting because it
is invisible from JSON and fatal to exactly the mechanism this paper depends on.
Croissant's context defines \code{conformsTo} as \code{dct:conformsTo} without
\code{"@type": "@id"} and sets \code{"@language": "en"} globally. The values of
the one property whose purpose is to identify a profile therefore expand to
language-tagged string literals rather than to IRIs: a conforming document
claims conformance to \emph{the text} \code{"https://w3id.org/croissant-policy/0.1.0"},
not to the profile that IRI names, so a profile identified through
\code{conformsTo} cannot be followed as an IRI from the expanded graph.

Three repairs exist and two are closed. Writing the value as a node reference
produces correct RDF and is rejected by \code{mlcroissant}, which
string-compares \code{conformsTo} to determine which Croissant version a
document claims and then fails it on every expectation that depended on the
version it could not determine; we made that change and reverted it. Adding a
typed term for \code{conformsTo} to the profile's own context would redefine a
Croissant core term, which the profile's additivity clause forbids. So
conforming documents keep the bare strings, the shapes accept either form, and
the defect is reported to MLCommons rather than absorbed silently---issue~1047,
with the reproduction~\cite{chernov2026conformsto}. It is a small thing with a
large consequence, and it is the kind of thing that is only found by running the
standard's own tooling against the standard's own output.

\subsection{The conformance corpus}
\label{sec:conformance}

Everything above is established on three documents. They are real, which is what
makes them the right evidence for the cost figures and for the claim that this
ran in front of something. They are three, which makes them weak evidence for a
different claim---that the translation is faithful and the carriers agree across
the documents the specification actually permits, rather than across the three
that happened to exist.

What is established has to be stated carefully, because the obvious
overstatement is available and wrong. The corpus does \emph{not} exhaust the
space of conforming documents and could not: operands are arbitrary numbers and
strings, an \code{in} set has any cardinality, an action carries any number of
conditions, a policy declares any number of actions. The space that is finite
and small is the set of things the specification \emph{names}---five operators,
three refusal classes, five conformance clauses, two carriers, and a handful of
structural forms. That set is enumerable, which is what makes deriving the
documents that exercise it possible rather than a matter of remembering to write
them. The evaluation already generates the request matrix from a descriptor
rather than listing requests by hand (Section~\ref{sec:c1}); the conformance
corpus applies the same move one level up.

\paragraph{Coverage, not count.} The corpus is 22 valid cases and 13 defect
cases. A valid case is stored in both carriers, so 22 cases are 44 documents; a
defect fixture is stored in the \code{cpol:} form only, so 13 cases are 13
documents, for 57 stored in total.

Eleven of the thirteen defects are additionally run through the ODRL carrier.
Two are not, and neither is a gap in the carrier. One declares two policies,
which has no ODRL form because the mapping is defined for exactly one policy
node. The other omits its conformance claim, which has no ODRL form because
translation \emph{writes} the claim---converting it produces a document carrying
a correct ODRL claim, so the defect cannot survive the conversion. That the
carrier enforces its own profile identifier is tested directly instead.

Those numbers are here for reproducibility and none of them is the result. Size
is not the coverage criterion: a larger corpus may exercise fewer specification
features than a smaller one built to exercise them all, and only the size is
visible from the size. What the corpus covers is:

\begin{center}
\begin{tabular}{@{}p{0.22\linewidth}p{0.70\linewidth}@{}}
\toprule
dimension & covered \\
\midrule
operators & all five \\[2pt]
verdicts & \code{PERMIT} and all three refusal classes \\[2pt]
clauses & 1 through 5, positively and negatively \\[2pt]
carriers & \code{cpol:} and ODRL \\[2pt]
shapes & single- and multi-condition actions, multi-action policies, lifecycle-state preconditions, wrong-typed operands, strip cases \\[2pt]
defects & unknown operator, absent and false \code{failClosed}, missing operand, missing condition name, duplicated condition name, more than one policy, a condition that is not a node, an unclaimed profile \\
\bottomrule
\end{tabular}
\end{center}

The coverage claim is checked twice, because a corpus that declares its own
coverage is asserting rather than showing. The cases carry tags; then the test
suite collects the verdicts and operators the corpus actually reaches when it is
run, and fails if a declared outcome is one no request produced.

\paragraph{What it establishes.} Coverage of the features in the table above is
established by the corpus as a whole, valid and defect cases together. The
three-way agreement result is narrower and is stated separately for that reason:
it is a property of the \emph{valid} cases, each expressed as a native
descriptor, as \code{cpol:} terms, and as an ODRL policy in \code{sc:usageInfo},
and decided over its generated request matrix. That is 184 requests per
representation and 552 complete decision records, all matching three ways. The
outcome mix is 20 permits, 83 condition violations, 44 undeclared actions and 37
state refusals, so the agreement is not established over a set of cases that
happens to be trivial. Every defect case refuses every request put to it: 91 requests through the
\code{cpol:} carrier across all 13, and 77 through the ODRL carrier across the
11 that have an ODRL form.

All 44 valid documents---both carriers---load in \code{mlcroissant} with zero
errors, and all 44 satisfy SHACL shapes. Which shapes is the part worth being
explicit about: a document is validated against the shapes of the profile it
\emph{claims}, and the two carriers claim different profiles. Running the
\code{cpol:} shapes over an ODRL document would report the absence of
\code{cpol:} terms in a document that never claimed to have any, which is not a
conformance result. So the profile publishes two shapes graphs and the validator
selects by the claim.

One difference between them is instructive rather than incidental. In the
\code{cpol:} graph the operator must be matched with \code{sh:pattern}, because
Croissant's global \code{"@language"} makes it a language-tagged literal and
\code{sh:in} compares RDF terms---\code{"cpol:min"@en} is not the term
\code{"cpol:min"}. In the ODRL graph the operator is a node reference, expands to
an IRI, and \code{sh:in} works directly. The carrier is the better RDF of the
two, which is worth saying given that the \code{cpol:} form is the normative
one.

Of the thirteen defects, eight are reported as shape violations. The remaining
five are invisible to SHACL, in two classes, and both are boundaries rather than
oversights.

One is a relationship between siblings. Two conditions declared under a single
name survive into the graph as two separate nodes, each individually satisfying
the condition shape; what is violated is that no two conditions of one action may
share a name. Expressing that requires a SHACL-SPARQL constraint, which is as
expressive as the query language and therefore the unbounded fragment
Section~\ref{sec:whynotshacl} declines everywhere else.

The other four are malformed operands, and they are invisible for a duller
reason: \code{cpol:expected} is typed \code{@json} in the profile's context, so
it expands to a single JSON literal and the shapes cannot see inside it.
Whether that operand is a list or a string is not a fact about the graph.

Both classes are refused by the evaluator and reported by the structural
validator, so nothing is unchecked; what the shapes cannot reach, something
else does.

\paragraph{What it found.} Five defects, none reachable from three well-formed
descriptors, which is the argument for having built it. Two came from generating
documents; three more came from asking what Table~\ref{tab:operators} actually
promises. Writing down a decision procedure turns out to be a good way to
discover that the implementation has a different one.

\code{cpol:in} with a numeric operand \emph{raised} rather than refusing, so a
malformed document crashed the evaluator instead of being refused by it. Worse,
\code{cpol:in} with a string operand fell through to substring matching, so a
policy written \code{\{"in": "illumina"\}}---forgetting the array, which is the
obvious authoring slip---silently \textbf{permitted} \code{"illu"}. A rule that
admits more than its author wrote is the failure this profile exists to prevent,
and it was reachable from a document no validator complained about. The repair
is the operand--observation distinction Table~\ref{tab:operators} now states:
a wrongly typed observation is an ordinary condition violation, a wrongly typed
operand is a defect in the policy and refuses at translation.

The third is the same shape and was the hardest to see, because the code looked
right. The evaluator compared with the implementation language's equality, and
in Python a boolean is a subclass of integer, so \texttt{True == 1}. A policy
written \code{equals: true} was therefore satisfied by an observed $1$, and
\code{in: [1, 2]} admitted \texttt{true}. Both admit a request the policy did not
describe. The comparison is now the one Table~\ref{tab:operators} states.
Replaying the deployment corpus under it altered none of its 46 decisions---the
three real descriptors carry no boolean conditions---so nothing measured or
published elsewhere depends on the change, and the paper is not quietly
restating an old number.

The ODRL carrier \emph{raised} where it should have refused. A condition that is
not a node---legal JSON, forbidden by the profile---was poisoned correctly by the
\code{cpol:} parser and crashed the ODRL emitter. A malformed document has to
produce a refusal and never an exception, or the two carriers stop being
equivalent exactly where equivalence matters.

Conformance clause 2 was validated and not enforced. A document could carry the
policy terms, omit the profile IRI from \code{conformsTo}, and still be decided.
That made the claim decorative, and left the two carriers inconsistent, because
the ODRL carrier already refuses an unrecognised \code{odrl:profile}---the
Information Model requires it, as Section~\ref{sec:whynotodrl} sets out. A
profile identifier that costs nothing to omit is not an identifier. The
\code{cpol:} carrier now refuses a document that does not claim the profile, and
the carriers agree.

\paragraph{What this corpus is not.} It is generated, so it says nothing about
which policies anyone writes. Any rate computed over it is a property of the
generator, and no figure in this subsection is a base rate. It also carries no
timing: performance is measured where the descriptors are real, which is the
other corpus.

\section{Composing independently governed authorities}
\label{sec:precedence}

\subsection{The question}

The question in this section is one of policy composition: what happens when a
single request falls under two authorities that are separately written,
separately administered, and attached to different things. That is the situation
whenever a policy travelling with a dataset meets a policy governing the agent
that asked, and it is the situation Rein and ACP put on one side of and agent
control planes on the other.

Concretely, an agent control plane holds a registry of agents with an owner and a
scope, evaluates each action against approved policy inline before execution, and
blocks violations. That is structurally the same mechanism as the gate above,
aimed at a different subject: it binds policy to the \emph{caller}, this profile
binds it to the \emph{data}.

Neither is complete, and the incompleteness is structural rather than
circumstantial. A caller-side rule cannot answer a question whose answer belongs
to the data, because the rule does not travel with the dataset. A data-side
descriptor cannot answer whether the caller is still in scope, because the
profile has no identity model and Section~\ref{sec:limits} says so.

\subsection{Method, and what is designed rather than measured}

The data side is the corpus above: three descriptors that gated a real run.
\textbf{The caller side is a model, and every number below inherits that.} Four
caller scopes are written in the structure the products describe---an identifier,
an assurance state, and entitlements conditioned on the request---representing a
broadly entitled agent whose own threshold is laxer than the dataset's, an agent
scoped to agree with the dataset, an inspection-only agent, and an agent whose
attestation has lapsed.

One deliberate control: both authorities are evaluated by the same function. A
caller scope is translated into the same descriptor model and handed to the same
\code{gate.authorize}. If the two halves used different evaluators, a
disagreement could come from the evaluators rather than from the policies, and
precedence would not be the only variable.

The caller is given the request context plus the dataset's \emph{name}. It is not
given the dataset's state, because that is exactly what does not travel with the
caller; supplying it would model a product that does not exist.

Five rules are enumerated rather than assumed: \code{deny-overrides} (permit only
if both permit), \code{caller-overrides}, \code{data-overrides}, and the two
single-sided deployments that actually exist today, \code{caller-only} and
\code{data-only}.

\subsection{The structural claim, stated before the counts}

The counts below are produced by caller scopes we wrote, so we state first the
part that does not depend on them.

\begin{quote}
\textbf{Observation 1 (incomparability).} Let $P_{\mathrm{caller}}$ and
$P_{\mathrm{data}}$ be the sets of requests admitted by a caller-side authority
and by a data-side descriptor. Then in general
\[
  P_{\mathrm{caller}} \not\subseteq P_{\mathrm{data}}
  \quad\text{and}\quad
  P_{\mathrm{data}} \not\subseteq P_{\mathrm{caller}},
\]
because the two authorities range over partially non-overlapping state spaces.
\end{quote}

The argument is about expressibility, not about which policies happen to be
written. A dataset's lifecycle state---\code{QC\_PASSED}, \code{PENDING},
\code{RETRACTED}---is a property of the data at a moment in time. It is not
derivable from caller identity or scope, so no caller-side policy can predicate
on it, and a caller-only deployment cannot refuse a request that is wrong solely
because of the state the data is in. Symmetrically, whether a caller's attestation
has lapsed, or which entitlements it holds, is a property of the caller. It is not
derivable from a dataset descriptor---which has no identity model at all,
Section~\ref{sec:limits}---so no data-side policy can predicate on it either.

Two authorities that can each express a class of rule the other cannot must, if
any such rule is written, admit requests the other refuses. Incomparability is
therefore a consequence of where the state lives, not of adversarially chosen
policies.

\subsection{An empirical witness}

The experiment below does not establish Observation~1; it witnesses it on a
concrete corpus and puts numbers on how far apart the two sides land for one
plausible set of scopes.

184 requests: 4 caller scopes $\times$ 3 datasets $\times$ the generated request
matrix.

\begin{center}
\begin{tabular}{lrr}
\toprule
agreement & count & share \\
\midrule
Both permit & 12 & 6.5\% \\
Both refuse & 137 & 74.5\% \\
Only the caller refuses & 12 & 6.5\% \\
Only the dataset refuses & 23 & 12.5\% \\
\bottomrule
\end{tabular}
\end{center}

\textbf{Neither permit set contains the other.} Both disagreement classes are
non-empty, so precedence is a real question rather than an artifact of one policy
being stricter throughout.

\begin{center}
\begin{tabular}{@{}p{0.42\linewidth}rr@{}}
\toprule
precedence rule & permits & admits despite a refusal \\
\midrule
\code{deny-overrides} & 12 & \textbf{0} \\[2pt]
\code{caller-overrides} / \code{caller-only} & 35 & \textbf{23} (12.5\%) \\[2pt]
\code{data-overrides} / \code{data-only} & 24 & \textbf{12} (6.5\%) \\
\bottomrule
\end{tabular}
\end{center}

The two rows that are not \code{deny-overrides} are the two products that exist.
An agent control plane operating alone admits every request in the fourth row of
the first table; an admission gate operating alone admits every request in the
third.

What each side structurally cannot reach is more interesting than the totals. Of
the 23 requests only the dataset refused, 20 violate a condition the caller never
carried and \textbf{3 turn on the dataset's lifecycle state}---a class of rule no
caller-side policy can express, because the state is not a property of the
caller. Of the 12 only the caller refused, 6 turn on the caller's own lapsed
assurance and 4 on an entitlement the caller does not hold: the mirror image, and
equally unreachable from a descriptor.

\paragraph{Cost.}
With both descriptors resident, the conjunction is $5.7\us$ against $2.5\us$ for
either side alone---the expected factor of two plus bookkeeping. Re-translating
both documents per decision costs $17.6\us$. Running both authorities is not
expensive.

\paragraph{One receipt.}
The joint record carries both verdicts, both refusal classes, both reason lists,
and both condition lists including the passing conditions of the authority that
permitted. A record that keeps only the refusing half cannot show the other
authority was consulted, which is the reason to have one record rather than two.

It also carries, for each of the five precedence rules, the verdict that rule
\emph{would have} yielded on the same request---not only the one applied. The
verdicts not reached are written into the record at decision time, so a later
reader determines from the record alone, without re-evaluating either policy and
without access to either document, whether the outcome depended on the precedence
rule in force. Where every rule agrees, the outcome is attributable to the
policies; where they differ, it is attributable to the configuration, and the
record says which. This matters because the policy that produced a decision is
frequently not retained in the form it had at the time, and a record that can only
be interpreted by re-running the system against a hypothetical configuration is
not evidence.

The same property makes the archive answerable after the fact. Because every
condition is stored with its observed value as well as its expected one, a stored
decision can be re-decided against a \emph{later} policy without the original
data, the original pipeline, or the evaluator that produced it---
\code{croissant\_policy/recheck.py} does this, and reports the fraction of an
archive that cannot be decided rather than guessing at it.

\subsection{What these numbers are not}

The rates are properties of the generated request matrix, which is deliberately
weighted toward violations---one per condition, plus wrong types and undeclared
actions. They are not base rates of production traffic, and the 19\% disagreement
figure should not be read as one.

The objection this invites is that four caller policies were written and they
turned out to be incomparable with the dataset's, which is unsurprising if they
were written to be. That objection applies to the counts and not to
Observation~1. The counts are a witness; the claim is that incomparability
follows from the two authorities ranging over different state spaces, and the
witness is informative because it shows the classes are non-empty in a corpus
nobody designed for the purpose---three of the dataset's refusals turn on
lifecycle state, six of the caller's on lapsed assurance and four on an
entitlement it does not hold.

The caller scopes are ours and a different set produces different counts. What a
different set cannot produce is a caller policy that predicates on dataset
lifecycle state, or a descriptor that predicates on caller assurance.

\section{What this does not do}
\label{sec:limits}

No identity and no entitlement: conditions are evaluated against a context, and
who supplied it is out of scope. No obligations or duties. No temporal or
stateful conditions---no windows, quotas or counters. No drift or relearning
semantics. Retention is carried and not applied, which the test suite asserts
rather than the prose merely stating.

Each omission is a place where the honest answer is that the mechanism is not
here, rather than a place where the profile quietly permits.

\section{Threats to validity}\label{sec:threats}

The \code{@context} has been checked against MLCommons'
\code{mlcroissant}~\cite{mlcommons2024croissantspec} 1.1.0:
all three example documents load with zero errors and the context is identical to
the one MLCommons' own generator produces for Croissant 1.0. Doing so found three
defects that assertion would not have. The hand-written context carried four
terms it should not have---two of them Croissant 1.1, two of them not Croissant
terms at any version---and was missing two; provenance of collection was being
written to a bare \code{dataCollection} key that resolved to a schema.org term
that does not exist, rather than through the \code{rai:} prefix; and the
decision-record node was a \code{cr:FileObject} without the checksum Croissant
requires, which the emitter now computes rather than invents. Two warnings remain
on the examples, both for recommended properties (\code{citeAs},
\code{datePublished}) that are supported as overrides and not fabricated.

That closes conformance clause 1 for the documents in \code{examples/}. The
remaining clauses are checked twice, and the two checks are not equivalent: the
profile's own validator inspects the JSON structurally, and the SHACL
shapes~\cite{knublauch2017shacl} of Section~\ref{sec:shapes} constrain the
expanded graph. The second is what caught the \code{conformsTo} defect, which
the first cannot see by construction.

One limit stays. The additivity clause is absent from the shapes and is not an
oversight: it is a statement about the document that \emph{remains} after every
profile term is deleted, and SHACL constrains the graph it is handed rather than
one derived from it by deletion. That clause is checked by performing the strip
and validating the result, which is a test rather than a shape.

The deployment corpus is three descriptors. They are real and they gated real
execution, and three is the number of them there are. The conformance corpus
(Section~\ref{sec:conformance}) answers the breadth question the deployment
corpus cannot, but it answers only that one: it is generated, so it is evidence
about the specification's feature space---the finite set of things the
specification names---and no evidence about production reality. Neither corpus substitutes for the other, and the paper does not report
a figure from one as though it came from the other.

The overhead figures are in-process and exclude interpreter startup, exactly as
the gate's own published figure does, which makes them comparable and makes both
an underestimate of what a per-task subprocess costs.

One identifier claim needs stating precisely, because on the Web a persistent
identifier that does not resolve is a promise rather than a fact. The profile IRI
is \url{https://w3id.org/croissant-policy/0.1.0}. The w3id.org registration that
makes it dereference was merged upstream on 2026-08-21, and the redirect was
verified end to end rather than assumed: a plain request returns
\code{303 See Other} to the served specification and then \code{200}, and the
same request under \code{Accept: application/ld+json} returns \code{303} to the
JSON-LD context and then \code{200} with a JSON-LD content type. Content
negotiation therefore works, and every \code{conformsTo} IRI a conforming
document emits resolves. An earlier version of this paper recorded the
registration as open and the IRI as not yet resolvable; that was true when
written and is no longer.

\section{Next}
\label{sec:next}

\begin{itemize}
\item \textbf{Replace the modelled caller side with a real one.} The structural
result above stands on a model of an agent control plane. Running the conjunction
against an actual product would convert the direction into a measurement. This is
the largest gap in the paper.
\item \textbf{A workload, not a matrix.} The disagreement rate needs a request
distribution that reflects something real before it means anything.
\item \textbf{Extend the deployment corpus.} Section~\ref{sec:conformance}
covers the specification's feature space; what it cannot supply is evidence
about production reality, and that needs more gates in front of more real pipelines
rather than more generated documents.
\item \textbf{Resolve the \code{conformsTo} defect upstream.} It is reported,
with the reproduction above, as MLCommons/croissant
issue~1047~\cite{chernov2026conformsto}. The fix is theirs to choose---the
context could type the term as \code{@id}, the reference implementation could
accept node references, or the global \code{"@language"} could be set to
\code{null}, which a third-party comment on the issue reports as a working
remedy. The issue remains open with no response from the project as of
2026-09-16. Until one of those lands, a Croissant profile identified
through \code{conformsTo} cannot be followed as an IRI from a descriptor's
expanded graph, which is a cost the whole format pays and not only this layer.
\item Take the profile to the MLCommons Data working group as a proposal. The
value of a profile is proportional to how many consumers implement it, and the
precedence result gives the working group a concrete reason to care: the profile
is the half of the pair their members' control planes cannot supply.
\end{itemize}

\section{Conclusion}

Croissant made datasets machine-readable, and version 1.1 made their conditions
of use machine-readable too. Neither made them answerable, because a condition
that can be written is not yet a request that can be decided. This paper adds
that half as an application profile rather than a replacement, in the pattern
DCAT-AP established over DCAT: a dataset declares the operations it admits and
the conditions under which it admits them, and a gate decides from the descriptor
alone, under an evaluation procedure the carrier does not supply. Keeping the operator set closed at five is what buys the
properties that matter---evaluation cost linear in the conditions written down,
an explanation in one line, and a refusal produced by the gate that was already
there rather than by a second decision engine. A more expressive vocabulary was
available and an ODRL profile could have carried the same conditions. We
established that by carrying them: the same policy expressed as an ODRL policy
in \code{sc:usageInfo}, where Croissant 1.1 says use conditions go, decides
every request in the matrix identically, record for record. Four of the five
operators turn out to be ODRL core operators. What the carrier does not bring
with it is the evaluation semantics those three properties rest on, because the
standard leaves that to whoever implements it---which is the paper's claim,
stated in the one form that can be checked rather than argued.

The evaluation is deliberately narrow and the claims are correspondingly small.
Decisions taken from a profile document match those taken from the native
descriptor of a gate that ran in front of a real nf-core pipeline, record for
record and not merely verdict for verdict. Stripping the layer leaves a document
MLCommons' own validator loads. The cost is $0\us$ warm and $11.7\us$ cold
against a $119\us$ decision, which is a disclosure rather than a claim of
negligibility---as Section~\ref{sec:threats} notes, the per-task subprocess that
delivers that decision costs more than two hundred times the decision itself,
and that is where a deployment should look first.

The result we did not expect is the precedence one. Agent control planes bind
policy to the caller and this profile binds it to the data, and over the same
request space neither permit set contains the other. That is not an argument that
one is better. It is evidence that a deployment consulting only one of them is
admitting requests some authority would have refused, in both directions, and
that the conjunction closes both gaps for roughly the cost of one extra
descriptor lookup. The caller side of that experiment is designed rather than
measured, and replacing it with a real control plane is the largest single
improvement available to this work.

\section*{Declaration of competing interests}

The author is employed by AstraZeneca. The work reports no AstraZeneca product,
system or dataset: the pipeline is the public \code{nf-core/demo}, the data is
public nf-core test data, and every artifact named in the Availability section
is published under an open licence. The employment is declared because it is
the kind of thing a reader is entitled to weigh, not because anything in the
paper turns on it.

\section*{Funding}

This research did not receive any specific grant from funding agencies in the
public, commercial, or not-for-profit sectors.

\section*{Availability}

The specification, reference implementation, conformance validator, SHACL
shapes, both carriers, the conformance-corpus generator with its manifest,
benchmarks and complete results are archived under
Apache-2.0 with a permanent DOI. The profile is identified by
\url{https://w3id.org/croissant-policy/0.1.0} and the ODRL carrier by that IRI
with \code{/odrl} appended; alongside them the deposit carries the profile's
\code{prof:Profile} description, which names each of those artifacts and the
role it plays. The w3id registration merged on 2026-08-21, so the profile IRI
dereferences---under content negotiation as well as plainly---and either it or
the DOI below may be used to retrieve the artifacts.

\medskip\noindent The archive is \href{https://doi.org/10.5281/zenodo.22018156}{doi:10.5281/zenodo.22018156},
a concept DOI that resolves to the current version of the deposit. Every result
in this paper is reproducible from v0.2.0, which is what it resolves to at the
time of writing.

\medskip\noindent The decision corpus the measured figures are taken from is
deposited separately under CC~BY~4.0 as
\href{https://doi.org/10.5281/zenodo.22016112}{doi:10.5281/zenodo.22016112}
\cite{chernov2026corpus}: 285 admission decisions from 30 interleaved
replicates of nf-core/demo v1.0.1, with the emitted Croissant descriptors and
per-file checksums.

\section*{Declaration of generative AI and AI-assisted technologies in the manuscript preparation process}

Assistance from generative AI tools (Anthropic Claude Opus~5) was used for
language review, for drafting and restructuring prose, and for code generation
in the reference implementation, its conformance test suite and the analyses
reported here. All content was reviewed, verified and edited by the author, who
takes full responsibility for the content of the published article.

\section*{CRediT authorship contribution statement}

\textbf{Alexander Chernov:} Conceptualization, Methodology, Software,
Validation, Formal analysis, Investigation, Data curation, Visualization,
Writing --- original draft, Writing --- review and editing.

\bibliographystyle{plain}
\bibliography{refs}

\end{document}